%% file: main.tex
\documentclass[runningheads]{llncs}

\include{def}

\usepackage{graphicx}

\usepackage{tikz}
\usepackage{comment}
\usepackage{amsmath,amssymb} 
\usepackage{color}

\usepackage[accsupp]{axessibility}  

\usepackage{booktabs} 
\usepackage{multirow}

\DeclareMathOperator*{\argmax}{arg\,max}

\usepackage{times}
\usepackage{epsfig}
\usepackage{graphicx}
\usepackage{amsmath}
\usepackage{amssymb}
\usepackage{enumitem}

\usepackage{xspace}
\usepackage{comment}
\usepackage{caption}

\usepackage{makecell}

\usepackage{booktabs}
\usepackage{xcolor,colortbl}

\usepackage{nicefrac}       
\usepackage{microtype}      
\usepackage[flushleft]{threeparttable}
\usepackage{multirow}
\usepackage{xcolor}
\usepackage{amsmath} 
\usepackage{bbm}
\usepackage{algorithm}
\usepackage{algorithmic}
\usepackage{amsfonts,amssymb}
\usepackage{booktabs,tabulary}
\usepackage{wrapfig}
\usepackage{multirow}
\usepackage{caption}
\usepackage{subcaption}
\usepackage{hyperref}

\definecolor{purplecolor}{RGB}{178,0,255}

\definecolor{LightCyan}{rgb}{0.95,1,1}

\begin{document}
\pagestyle{headings}
\mainmatter
\def\ECCVSubNumber{1791}  

\title{Improving Robustness by Enhancing Weak Subnets} 

\titlerunning{Enhancing Weak Subnets}
%
\author{Yong Guo \and
David Stutz \and
Bernt Schiele}


\authorrunning{Y. Guo et al.}
%
\institute{Max Planck Institute for Informatics, Saarland Informatics Campus, Saarbrücken
\email{\{yongguo,david.stutz,schiele\}@mpi-inf.mpg.de}}


\maketitle

\begin{abstract}
    Despite their success, deep networks have been shown to be highly susceptible to perturbations, often causing significant drops in accuracy. In this paper, we investigate model robustness on perturbed inputs by studying the performance of internal sub-networks (subnets). Interestingly, we observe that most subnets show particularly poor robustness against perturbations. More importantly, these weak subnets are correlated with the overall lack of robustness. Tackling this phenomenon, we propose a new training procedure that \textbf{identifies and \underline{e}nhances \underline{w}eak \underline{s}ubnets (EWS) to improve robustness}. Specifically, we develop a search algorithm to find particularly weak subnets and explicitly strengthen them via knowledge distillation from the full network. We show that EWS greatly improves both robustness against corrupted images as well as  accuracy on clean data. Being complementary to popular data augmentation methods, EWS consistently improves robustness when combined with these approaches. To highlight the flexibility of our approach, we combine EWS also with popular adversarial training methods resulting in improved adversarial robustness.
\end{abstract}

\section{Introduction}
\label{sec:intro}

Since 2012, when AlexNet won the first place in the ImageNet competition~\cite{krizhevsky2012imagenet}, deep (convolutional) networks~\cite{lecun1989backpropagation} have been producing state-of-the-art results in many challenging tasks~\cite{he2015deep,Lee2015}.
Recent work, however, highlights how brittle these models are when applied to images with simple corruptions, such as noise, blur, and pixelation~\cite{HendrycksICLR2019}. In fact, these corruptions cannot fool the human vision system but often severely hamper the accuracy of deep networks~\cite{HendrycksICLR2019,hendrycks2021many}.
Among an increasing body of work on developing robust models, data augmentation is particularly popular and effective.
For example, AutoAugment \cite{cubuk2018autoaugment}, AugMix \cite{hendrycks2019augmix} or DeepAugment \cite{hendrycks2021many}
improve the robustness against corrupted examples alongside the clean accuracy.

\begin{figure}[t!]
    \centering
    \includegraphics[width = 1.0\textwidth]{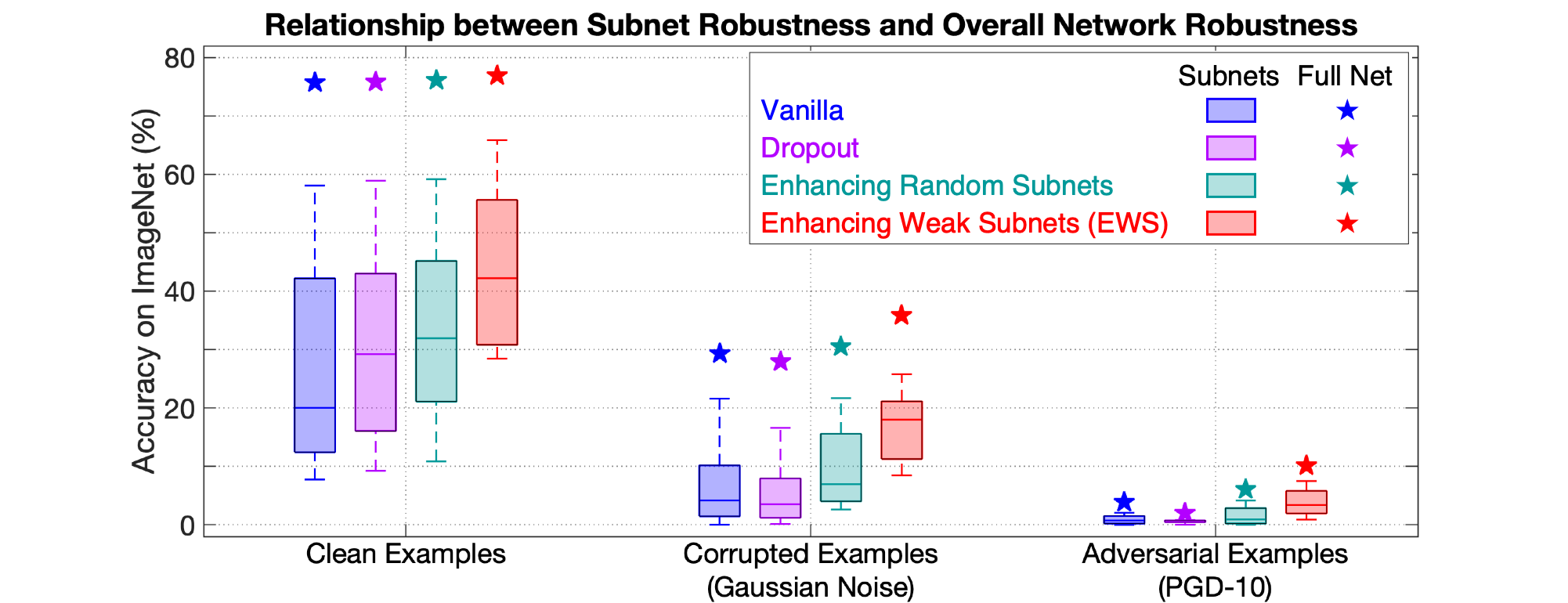} 
	\caption{On ImageNet, we plot accuracies on clean and perturbed examples for a standard ResNet-50 (\textcolor{blue}{blue stars}) and 1K randomly sampled subnets (\textcolor{blue}{blue box plot}) corresponding to $70\%$ of the paths/channels each.
	On clean examples (left), subnets perform significantly worse than the overall network, with average accuracy reducing to roughly $20\%$.
	On examples perturbed with Gaussian (middle) or adversarial (right) noise, overall accuracy reduces severely alongside subnet accuracy, suggesting that \emph{weak subnets} are responsible for this lack of robustness.
	By identifying and \underline{e}nhancing these \underline{w}eak \underline{s}ubnets (EWS, \textcolor{red}{red}), we can improve robustness significantly.
	Importantly, we search for particularly weak subnets in order to enhance them through knowledge distillation, while encouraging distributed representations through dropout \cite{srivastava2014dropout} (\textcolor{violet}{violet}) or improving \emph{random} subnets (\textcolor{green!50!black}{green}) does not improve accuracy or robustness as much. 
	}
	\label{fig:motivation}
\end{figure}

Complementary to this line of research, we study the robustness of nowadays over-parameterized networks by analyzing their internal sub-networks (subnets).
While it is well-known that few well-performing subnets, \ie, ``winning tickets'', exist within these large networks~\cite{FrankleARXIV2018,SehwagARXIV2020,LiARXIV2020}, the role of the remaining subnets in terms of robustness remains unexplored.
In this paper, we find that most of these subnets perform rather poorly.
Moreover, this is strongly correlated with the network's overall performance and becomes particularly apparent on corrupted or adversarial examples.
As illustrated in Fig.~\ref{fig:motivation}, for a standard ResNet-50 \cite{he2015deep} on ImageNet \cite{russakovsky2015imagenet}, 
the network's accuracy (\textcolor{blue}{blur star}) reduces significantly alongside the severely degraded subnets' accuracy (\textcolor{blue}{blue box}) when facing perturbed examples, \eg, with Gaussian noise.
Note that the \textcolor{blue}{blue box} illustrates the performance range of 1K randomly sampled subnets, each corresponding to $70\%$ of the overall network, and the mean performance (\textcolor{blue}{blue line in box}) is at the lower end of this range.
Overall, this leads us to the hypothesis that these \emph{weak subnets} are, to some degree, responsible for the lack of robustness of the overall network.

In order to test this hypothesis and improve robustness, 
we intend to enhance subnet performance.
Interestingly, dropout~\cite{srivastava2014dropout} can be regarded as a particular approach to construct and train subnets by randomly dropping the internal connections.
Unfortunately, as shown in Fig.~\ref{fig:motivation}, it only slightly improves the subnets on clean data but yields much worse performance on perturbed data.
It is worth noting that, the correlation between subnet and full network still holds, \ie, better subnets (\textcolor{blue}{blue boxes} $>$ \textcolor{violet}{violet boxes}) always come along with better full network (\textcolor{blue}{blue stars} $>$ \textcolor{violet}{violet stars}) on perturbed data (middle and right of Fig.~\ref{fig:motivation}). Thus, regarding this correlation, how to effectively improve subnets becomes an important problem to boost the overall robustness against perturbations.

\textbf{Contributions:}
To address this, we propose to directly and very explicitly identify and improve weak subnets instead of the random ones, which is proved to be a more effective way (see Fig.~\ref{fig:motivation} and Table~\ref{tab:comparison_search}).
Specifically, we make three key contributions:
\emph{(1)} We propose a novel robust training method which identifies and \textbf{enhances weak subnets (EWS)} to improve the overall robustness of the full network.
\emph{(2)} To this end, we develop a search algorithm that obtains weak subnets by identifying particularly weak paths/channels inside the full network.
Given a weak subnet, its performance is further enhanced by distilling knowledge from the full network.
This approach is not only very scalable, it also adds negligible computational overhead (see results in Section~\ref{subsec:ablation}).
Note that identifying particularly weak subnets is crucial as shown in Fig.~\ref{fig:motivation} where enhancing \emph{random} subnets has little effect (\textcolor{red}{red stars and boxes} $>$ \textcolor{green!50!black}{green stars and boxes}).
\emph{(3)} In experiments, we apply EWS on top of state-of-the-art data augmentation schemes to improve accuracy and corruption robustness on CIFAR-10/100-C and ImageNet-C \cite{hendrycks2021many}.
Moreover, we also demonstrate the generality of our approach for improving adversarial robustness on top of recent adversarial training methods.
Importantly, our approach is complementary to all these methods and improves consistently across a wide range of approaches.
Our code is available at \href{https://github.com/guoyongcs/EWS}{https://github.com/guoyongcs/EWS}.

\section{Related Work}
\label{sec:related}

Despite their outstanding performance, deep networks are not robust to image corruptions ~\cite{hendrycks2021many}. 
To address this, recent work explores re-calibrating batch normalization statistics \cite{SchneiderARXIV2020,BenzARXIV2020,NadoARXIV2020}, utilizing the frequency domain \cite{SaikiaARXIV2021}, 
or using vision transformers \cite{MaoARXIV2021} to improve corruption robustness.
However, data augmentation methods such as \cite{RusakECCV2020,geirhos2018imagenet,cubuk2018autoaugment,hendrycks2019augmix,hendrycks2021many,CalianARXIV2021,SalmanARXIV2020} represent the most prominent and successful line of work, ranging from simple Gaussian noise augmentation \cite{RusakECCV2020}, over well-known schemes such as AutoAugment \cite{cubuk2018autoaugment} to strategies specifically targeted towards corruption robustness such as AugMix \cite{hendrycks2019augmix} or DeepAugment \cite{hendrycks2021many}.

Besides random corruptions, deep networks are susceptible to adversarial examples \cite{SzegedyARXIV2013,GoodfellowARXIV2014}.
While plenty of approaches for defending against adversarial examples have been proposed~\cite{SilvaARXIV2020,BarrenoASIACCS2006,YuanARXIV2017,AkhtarARXIV2018,BiggioARXIV2018,XuARXIV2019,ChaubeyARXIV2020,rebuffi2021fixing,PangICLR2021,GowalARXIV2020}, adversarial training (AT) has become the de facto standard \cite{madry2017towards}.
This is also because other methods have repeatedly been ``broken'' using stronger or adaptive attacks, \eg, see \cite{CarliniARXIV2016,CarliniARXIV2017,CarliniARXIV2017b,LiuARXIV2018,EngstromARXIV2018,MosbachARXIV2018,AthalyeARXIV2018b,CarliniARXIV2019b,CroceICML2020}.
Thus, recent work concentrates on various variants of AT:
\cite{ZhangICML2019} adds an additional Kullback-Leibler term, \cite{HendrycksNIPS2019} incorporates a complementary self-supervised loss, \cite{UesatoNIPS2019,CarmonNIPS2019} use additional unlabeled examples. Many more variants exploring, \eg, instance-adaptive threat models \cite{BalajiARXIV2019,DingICLR2020}, additional regularizers \cite{ZhangNIPS2019,RakinCVPR2019,PangNIPS2020,PangNIPS2020,SinglaICML2020,WanECCV2020,BuiECCV2020,LiICCV2021}, 
curriculum training \cite{CaiIJCAI2018,YuARXIV2019},
weight perturbations~\cite{wu2020adversarial,StutzICCV2021}, 
among many others \cite{LambAISEC2019,WangNIPS2020,ChengARXIV2020,ZhangICLR2019,ZhangICML2020,JeddiARXIV2020,ZiICCV2021}, have been proposed.
Moreover, AT has also been applied to corruption robustness \cite{KireevARXIV2021,MadaanARXIV2020}. 

Complementary to the above lines of research, existing work~\cite{LiARXIV2020,SehwagNIPS2020,diffenderfer2021winning} have shown the existence of particularly strong/robust subnets inside a large model. However, the impact of the remaining weak subnets on the overall robustness has not been investigated. 
Interestingly, as shown by Fig.~\ref{fig:motivation}, we observe that the majority of subnets are particularly weak and there is a clear correlation between the performance of subnets and overall robustness. This motivates us to explicitly enhance these weak subnets.
To achieve this goal, our method is also inspired by recent work on neural architecture search (NAS)~\cite{zoph2016neural,liu2018darts,cai2018proxylessnas,chen2021contrastive,guo2021towards} and knowledge distillation (KD)~\cite{hinton2015distilling,zi2021revisiting}.
Unlike existing NAS methods, we focus on improving the robustness of deep models and exploit NAS techniques to find weak subnets which may hamper the overall performance.
Once we find these weak subnets, we seek to further enhance them using a KD loss, similar to learning lightweight student models in model compression \cite{sun2019patient,liu2020discrimination,gou2021knowledge}.

\section{EWS: Training by \underline{E}nhancing \underline{W}eak \underline{S}ubnets}
\label{sec:method}

This paper studies the robustness of deep, over-parameterized networks to image perturbations through the lens of subnets defined by a subselection of internal blocks/channels.
After arguing that the robustness of subnets has a large impact on overall performance in Section \ref{subsec:subnets}, we seek to improve robustness through a novel training procedure:
identifying and enhancing weak subnets through knowledge distillation, as summarized in Fig.~\ref{fig:overview}.
Specifically, in Section \ref{subsec:search}, we first discuss the construction and search of particularly weak subnets along with the motivation for strengthening them.
Here, we present an effective and scalable search algorithm to find weak subnets.
In Section \ref{subsec:training}, we then integrate the found weak subnets into our training procedure to explicitly enhance them.
Our procedure is summarized in Algorithm \ref{alg:training}.
In this paper, we mainly concentrate on improving robustness to corrupted images, \eg, with noise or blur, on standard benchmarks such as CIFAR-10-C and ImageNet-C~\cite{hendrycks2021many}.
Besides, we also highlight the flexibility of our method by applying it on top of adversarial training \cite{madry2017towards}.

\subsection{Subnet Construction and Impact on Overall Performance}
\label{subsec:subnets}

\begin{figure}[t]
    \centering
    \includegraphics[width=0.7\textwidth]{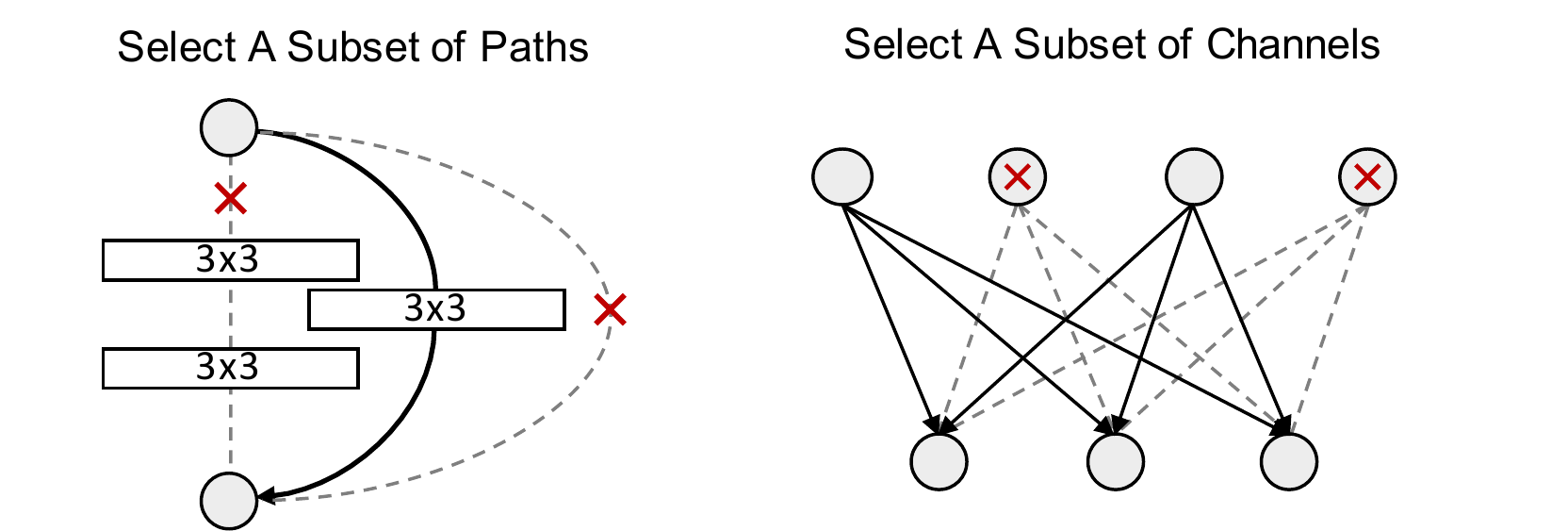}
	\caption{
	We construct subnets by selecting a subset of paths (left) and channels (right) in each block of the network.
	Left: An example for a basic block with multiple paths 
	where only one path is kept (in \textbf{bold}).
	Right: In the convolutional (fully connected) layers of each block, we select a subset of channels (in \textbf{bold}).
	This approach is applied block-by-block, layer-by-layer using a fixed fraction of paths/channels to be selected.
	}
	\label{fig:subnet}
\end{figure}

We intend to investigate the robustness of deep networks from the perspective of subnets.
To this end, we define subnets as follows:
Given a full (convolutional feedforward) network $M$, we construct a set of subnets $\alpha \in \Omega$ where $\Omega$ denotes the space of all the possible subnets of $M$.
In this paper, we particularly consider deep models that consist of a stack of basic blocks, \eg, ResNets~\cite{he2015deep}. 
Actually, our approach can easily be adapted to any other architectures since most popular designs (\eg, MobileNet~\cite{HowardARXIV2017,sandler2018mobilenetv2} and ResNeXt~\cite{xie2017aggregated}) can be also decomposed into the basic units defined in Fig.~\ref{fig:subnet}.
Without loss of generality, we allow each basic block $i$ to have a specific number of paths $n_i$ and
each layer $j$ in a block to have a specific number of channels $c_j$.
For example, a basic residual block~\cite{he2015deep} contains $n_i = 2$ paths (\ie, the residual path and the skip connection).
As shown in Fig.~\ref{fig:subnet},
we construct subnets by selecting a subset of paths and channels in each block and layer, respectively.
Given a subnet width $\rho \in (0,1)$, we select roughly $\rho \cdot n_i$ paths in each block $i$ and $\rho \cdot c_j$ channels for each layer $j$.
For example, $\rho = 0.7$ denotes selecting 70\% of paths and channels.

We further investigate the impact of subnets on the overall performance.
In Fig.~\ref{fig:motivation}, we quantify the robustness of such subnets ($\rho {=} 0.7$) on corrupted and adversarial examples:
Even on \emph{clean} examples (left) where the full network tends to make correct predictions with relatively high accuracy (\textcolor{blue}{blue star}), more than 50\% of the subnets yield an accuracy below 20\% (\textcolor{blue}{blue box}).
On corrupted or adversarial examples (middle and right), this phenomenon is further emphasized.
More critically, as the performance of subnets deteriorates on such perturbed examples, the network's overall performance also reduces significantly.
{It is worth noting that, each subnet can be regarded as an indicator to show whether the corresponding parameters inside the full model are well learned or not. In this sense, one may easily improve the overall performance if we put more focus on these ``weak'' parameters during training.}
This motivates us to explicitly find and enhance these weak subnets.
However, we emphasize that merely training random subnets,
\eg, dropout \cite{srivastava2014dropout} (\textcolor{violet}{violet} in Fig.~\ref{fig:motivation}), does \emph{not} contribute to the improved robustness.

\subsection{Finding Particularly Weak Subnets}
\label{subsec:search}

In order to improve subnet performance and following Fig.~\ref{fig:overview}, our proposed EWS has two core components: finding particular weak subnets and strengthening their performance.
Regarding the first component (Fig.~\ref{fig:overview}, left), given a search space $\Omega$,
the easiest way to identify/enhance such subnets is by random sampling, \ie, randomly selecting blocks and channels for a given $\rho$.
However, Fig.~\ref{fig:motivation} (\textcolor{green!50!black}{green}) shows that improving the performance of random subnets during training (as described in detail in Section \ref{subsec:training}) does not result in significantly more robust subnets (\textcolor{green!50!black}{box plot}) or improved overall performance (\textcolor{green!50!black}{star}).
Instead, we propose a novel search algorithm that finds particularly poor subnets $\alpha$, as evaluated based on their classification accuracy $R(\alpha)$.
Following~\cite{pham2018efficient,guo2020breaking}, the accuracy can be approximated based on a batch of sampled data in each iteration.
To be specific, we learn a policy $\pi_{\theta}$, parameterized by $\theta$, and use it to generate candidates of weak subnets, \ie, $\alpha \sim \pi_{\theta}$, as identified based on particularly low accuracy.
To learn the policy, we build a controller model to produce candidate subnets by minimizing the accuracy $R(\alpha)$ in expectation.
Formally, we solve the following optimization problem
\begin{equation} \label{eq:search}
\begin{aligned}
&\min_{\theta}~\mathbb{E}_{\alpha\sim \pi_{\theta}} \left[ R(\alpha)  \right].
\end{aligned}
\end{equation}
The controller's parameters can be updated using policy gradient based on a mini-batch of sampled subnets (see the supplementary material for details). This is made explicit in the first part (Lines 3-7) of Algorithm~\ref{alg:training}.

\begin{figure*}[t]
    \centering
    \includegraphics[width = 1.0\textwidth]{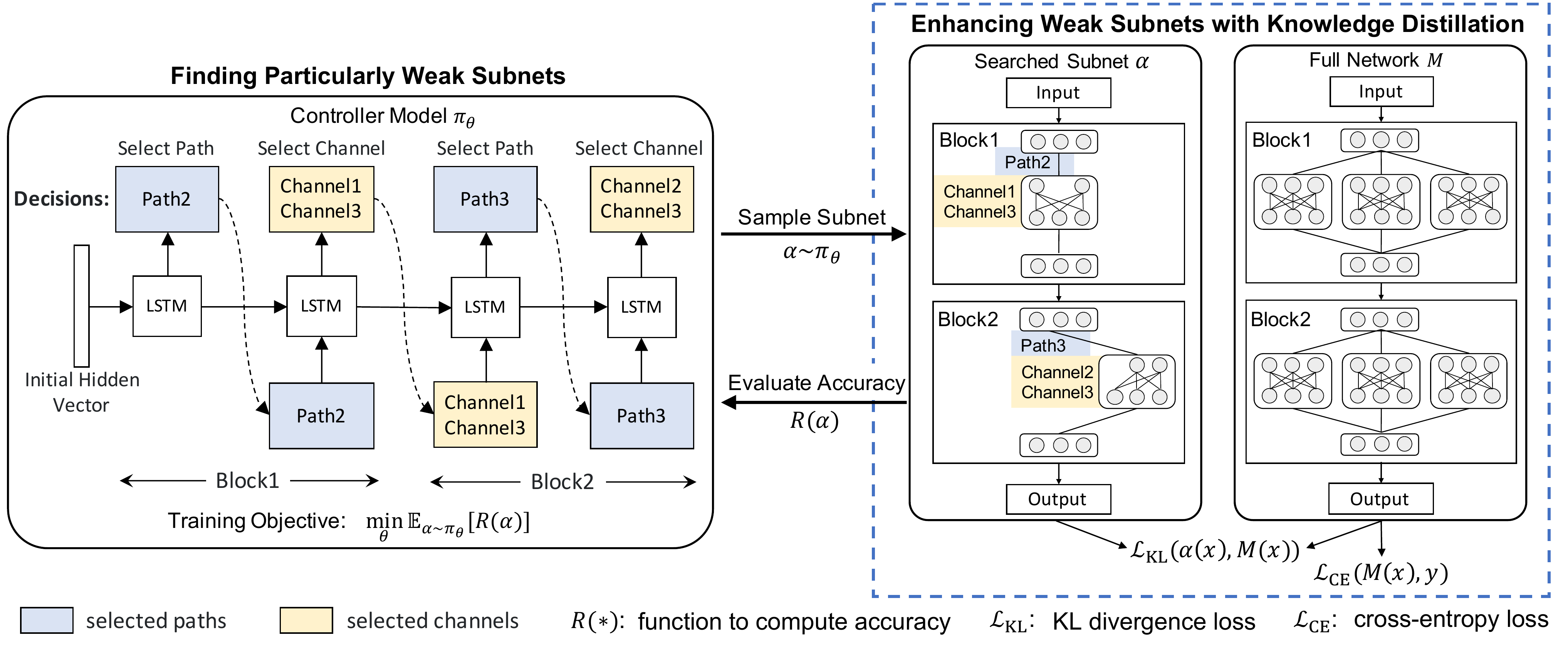}
	\caption{
	Overview of our proposed \textbf{enhancing weak subnets (EWS)} training procedure.
	During training, we alternatingly perform subnet search based on a controller model (left) and train the full network with an additional distillation loss (right).
	As illustrated, the controller is 
	trained using policy gradient every $K$ iterations,
	while the full network can be any state-of-the-art network such as ResNets \cite{he2015deep}.
	The distillation loss enforces predictions of the full network and the found weak subnet to be similar to improve subnet performance. 
	We refer to Algorithm \ref{alg:training} for a detailed description.}
	\label{fig:overview}
\end{figure*}

\textbf{Architecture of the Controller Model.}
Since a network like ResNet can be represented by a series of tokens~\cite{pham2018efficient}, the subnet generation task can be viewed as a sequential decision making problem.
Following~\cite{pham2018efficient,guo2020breaking}, we build an LSTM-based~\cite{graves2012long} controller model to learn the policy $\pi_{\theta}$.
Specifically, the controller takes an initial hidden state (kept constant during training) as input and
sequentially predicts the selected paths/channels for each block/layer, as illustrated exemplarily in Fig.~\ref{fig:overview} (left).
Specifically, for each block, we first select a subset of paths and subsequently select a subset of channels for each layer in the selected paths.
We emphasize that the controller model scales linearly with the overall network and is easily applicable to a wide range of architectures.
Please refer to our supplementary for further details.

\subsection{EWS: \underline{E}nhancing \underline{W}eak \underline{S}ubnets with Knowledge Distillation}
\label{subsec:training}

\begin{algorithm}[t]
\caption{Training by \textbf{enhancing weak subnets (EWS)}: We alternate between updating the controller model $\pi_{\theta}$ (every $K$ iterations) and updating the model $M$. 
When training $M$, we sample a subnet $\alpha {\sim} \pi_{\theta}$ and exploit a distillation loss to enhance it.
}
\label{alg:training}
    \begin{algorithmic}[1]
    	\REQUIRE
    	Training data $\mD$, batch size for training model $N$, batch size for training controller $C$, hyper-parameter $\lambda$, training interval $K$, number of iterations $T$\\
        \FOR{$t=1, \cdots, T$}
        \STATE Sample a batch of examples $\mathbb{B} = \{(x_i, y_i)\}_{i=1}^{N}$ from $\mD$
        \STATE // \emph{Update the controller every $K$ iterations}
        \IF{$t \bmod K=0$}
            \STATE Sample a set of subnets $\{\alpha_i\}_{i=1}^C$ from $\pi_{\theta}$
            \STATE Compute subnet accuracy $R(\alpha)$ on $\mathbb{B}$
            \STATE Update $\theta$ using policy gradient:\\
           ~~~~~~~~$\theta \leftarrow \theta + \eta \frac{1}{C} \sum_{i=1}^C \nabla_{\theta} \log \pi_{\theta}(\alpha_i) R(\alpha_i)$
        \ENDIF
        \STATE // \emph{Train the full model while enhancing weak subnets}
        \STATE Sample a weak subnet $\alpha \sim \pi_{\theta}$
        \STATE Update $w$ using gradient descent: \\
        ~~~~~~~~$w \leftarrow w - \eta \frac{1}{N} \sum_{i=1}^N \big( \nabla_{w} \mL_{\rm CE}(M(x_i),y_i) + \lambda \nabla_{w} \mL_{\rm KL}(\alpha(x_i), M(x_i)) \big)$
        \ENDFOR
    \end{algorithmic}
\end{algorithm}

In the second step of EWS, corresponding to Fig.~\ref{fig:overview} (right) and based on the found weak subnets $\alpha \sim \pi_{\theta}$, we use a distillation loss to enhance subnet performance and thereby aim to improve the overall robustness of the full network.
Let $x$ be a training example with its label $y$. 
Besides the standard cross-entropy (CE) loss, we introduce an additional Kullback-Leibler (KL) divergence loss between the full network's predictions $M(x)$ and the subnet's predictions $\alpha(x)$.
This can be thought of as distilling knowledge from the full network into the subnet and is meant to 
enhance the selected particularly weak subnets.
Overall, the loss function to be minimized becomes
\begin{equation}
\small
\begin{aligned}
  \mL(x,y) = \mL_{\rm CE}\big(M(x),y \big) + \underbrace{\lambda \mL_{\rm KL}\big(\alpha(x), M(x) \big)}_{\rm enhance~weak~subnet},
\end{aligned}
\label{eq:ews_clean}
\end{equation}
where $\lambda$ is a trade-off parameter determining the importance of enhancing weak subnets (see impact of $\lambda$ in Section~\ref{subsec:ablation}).
While other losses for improving weak subnets are possible, \eg, a CE loss with the true labels $\mL_{\rm CE}\big(\alpha(x), y \big)$, we found that distillation with the KL divergence works best in practice (see supplementary for details).
We emphasize that our method is very flexible in that it can easily be combined with different losses and is entirely complementary to data augmentation approaches.

As indicated in Algorithm~\ref{alg:training}, it is worth noting that, once we update the model parameters, the previously learned controller $\pi_\theta$ is outdated.
This is problematic as the sampled subnets $\alpha \sim \pi_\theta$ might not be particularly weak anymore.
Therefore, the model and controller are updated in an alternating fashion.
However, we found that it is not necessary to update the controller model in each iteration.
Instead, we only update the controller every $K$ iterations where $K = 1$ represents fully alternating training, while $K \gg 1$ updates the controller more rarely.
This also reduces the computational overhead of our methods significantly.
As discussed in detail in Section~\ref{subsec:ablation}, we find that $K = 10$ works very well in practice and leads to a negligible computational overhead.

\subsection{Combining EWS with Adversarial Training}
\label{subsec:adversarial}

We further demonstrate the flexibility of EWS by combining it with adversarial training to improve adversarial robustness.
Instead of identifying weak subnets by accuracy, we now consider 
\emph{robust} accuracy on adversarial examples.
Similarly, we enhance these subnets using a distillation loss on adversarial examples instead of the clean ones.

\textbf{Searching for \emph{Adversarially} Weak Subnets.}
Here, we seek to find subnets that are vulnerable to adversarial examples.
We follow \cite{madry2017towards} and compute adversarial $L_\infty$ perturbations by maximizing cross-entropy loss.
Specifically, we construct the perturbed data as $x'=\argmax_{\| x - x' \|_p \leq \epsilon} \mathcal{L}_{\text{CE}}(M(x'),y)$.
While other objectives are possible, this is commonly achieved using projected gradient ascent (PGD), where the projection reduces to a simple clamping operation for ensuring $\|x - x'\| \leq \epsilon$.
Finally, to find adversarially vulnerable subnets, we replace the (clean) accuracy in Equation \eqref{eq:search} with the adversarial accuracy.
We also note that the adversarial examples are computed with respect to the full model $M$ and re-computed every iteration.
Again, we follow Algorithm~\ref{alg:training} and train the controller using policy gradient.

\textbf{Adversarially Enhancing Weak Subnets.}
With the found weak subnets, we again enhance them to improve the overall adversarial robustness.
Following the vanilla adversarial training~\cite{madry2017towards}, we optimize the cross-entropy loss on adversarial examples:
\begin{equation}
\min_{w}~ \mathbb{E}_{(x,y) \sim \mD} \left[ \mL(x',y)\right]\quad{\rm where~} x' {=} \argmax_{\| x - x' \|_p \leq \epsilon} \mL_{\rm CE}(M(x'), y).\label{eq:adv_loss}
\end{equation}
Again, we additionally introduce a distillation-based KL divergence loss to enhance the subnets, following Section \ref{subsec:training} above.
Formally, the loss in Equation \eqref{eq:adv_loss} becomes
\begin{equation}
\mL(x',y) = \mL_{\rm CE}(M(x'), y) + \underbrace{\lambda \mL_{\rm KL}\big(\alpha(x'), M(x') \big)}_{\rm enhance~subnet~on~x'}\\ \\
\end{equation}
Note that the KL divergence is computed on the predictions $M(x')$ and $\alpha(x')$ based on the adversarial examples $x'$.
As before, $\lambda$ determines the importance of weak subnet performance.
Even though this formulation follows vanilla adversarial training, our approach generalizes easily to other variants, including TRADES \cite{zhang2019theoretically} where an additional loss on clean examples is minimized. In such cases, the KL divergence used for distillation is also computed on clean examples (see supplementary for more details).
Additionally applying weight perturbations \cite{wu2020adversarial} or using additional unlabeled examples \cite{CarmonNIPS2019} is possible as well (see Table~\ref{tab:adv_resnet}).
While Fig.~\ref{fig:motivation} only considers a model trained without adversarial examples, \ie, following Section \ref{subsec:training}, enhancing subnets during training clearly has a positive impact on robustness against adversarial examples (\eg, PGD ones as computed in \cite{madry2017towards}).
Thus, we expect a similarly positive impact when explicitly finding and enhancing adversarially vulnerable subnets during adversarial training.

\section{Experiments}
\label{sec:experiments}

In the following, we demonstrate that EWS allows to train more accurate and robust models, as tested mainly against corrupted examples in Section \ref{subsec:experiments-corrupted}.
Besides the standard training settings, we emphasize that EWS can be successfully used in a complementary fashion on top of popular data augmentation schemes, including AutoAugment \cite{cubuk2018autoaugment}, AugMix \cite{hendrycks2019augmix} or DeepAugment \cite{hendrycks2021many}.
Moreover, we also show that EWS easily generalizes to adversarial training without significant modifications and greatly improves the adversarial robustness in Section \ref{subsec:experiments-adversarial}. 
We provide additional ablations in Section \ref{sec:ablation}.

\subsection{Improving Corruption Robustness}
\label{subsec:experiments-corrupted}

We start by training our EWS on standard benchmark datasets, \ie, CIFAR-10~\cite{krizhevsky2009learning}, CIFAR-100~\cite{krizhevsky2009learning}, and ImageNet~\cite{deng2009imagenet}, and testing on both clean test examples and the corrupted ones, namely CIFAR-10-C, CIFAR-100-C, and ImageNet-C.
On CIFAR-10-C and CIFAR-100-C, we report the test error on clean or corrupted examples.
On ImageNet-C, in contrast, we report the mean corruption error (mCE)~\cite{HendrycksICLR2019}.
We also consider ImageNet-P which evaluates the prediction stability on videos using mean flip rate (mFR).
For all the metrics, \emph{lower is better}.
In all the experiments, by default, we set $K=10$, $\lambda=1$, and $\rho=0.7$ for our EWS.

\begin{table}[t]
  \centering
  \resizebox{1.0\textwidth}{!}
  {
    \begin{tabular}{c|c|c|>{\columncolor{LightCyan}}c|c|>{\columncolor{LightCyan}}c}
    \toprule
    \multicolumn{2}{c|}{\multirow{2}[0]{*}{Method}} & \multicolumn{2}{c|}{CIFAR-10} & \multicolumn{2}{c}{CIFAR-100} \\
    \cmidrule{3-6}
    \multicolumn{2}{c|}{} & \multicolumn{1}{c|}{~~Clean Error (\%) $\downarrow$~~} & {~~Corruption Error (\%) $\downarrow$~~} & \multicolumn{1}{c|}{~~Clean Error (\%) $\downarrow$~~} & {~~Corruption Error (\%) $\downarrow$~~} \\
    \hline
    \multirow{3}[0]{*}{Standard}  & Vanilla & 5.32 (-0.00)  & 26.46 (-0.00) & 23.45 (-0.00) &	50.76 (-0.00) \\
    & ~Dropout~ & 5.16 (-0.16)  & 26.17 (-0.29) & 23.19 (-0.26) &	50.43 (-0.33) \\
    & EWS &	\textbf{4.44 (-0.88)} &	\textbf{24.94 (-1.52)} & \textbf{22.41 (-1.04)} &	\textbf{40.08 (-1.68)} \\
    \hline
    \multirow{3}[0]{*}{~AutoAugment~} & Vanilla & 4.05 (-0.00)	& 16.19 (-0.00) & 23.02 (-0.00) & 44.37 (-0.00)
     \\
     & Dropout & 3.91 (-0.14) & 16.04 (-0.15) & 22.84 (-0.18) & 44.09 (-0.28) \\
     & EWS & \textbf{3.23 (-0.82)} &	\textbf{14.31 (-1.88)} & \textbf{22.16 (-0.86)} & \textbf{42.40 (-1.97)} \\
    \hline
    \multirow{3}[0]{*}{AugMix} & Vanilla & 4.35 (-0.00) & 13.57 (-0.00) & 22.45  (-0.00) & 38.28 (-0.00) \\
    & Dropout & 4.19 (-0.16) & 13.44 (-0.13) & 22.11 (-0.34) & 37.97 (-0.31) \\
    & EWS & \textbf{3.76 (-0.59)}  & \textbf{10.80 (-2.77)} & \textbf{21.81 (-0.64)} & \textbf{35.24 (-3.04)} \\
    \toprule
    \end{tabular}%
    }
  \caption{Clean and corrupted test error on CIFAR-10(-C) and CIFAR-100(-C). Our proposed EWS approach not only improves clean test error, but also consistently reduces corrupted test error as highlighted in \textbf{bold} on top of different data augmentation schemes.
  }
  \label{tab:cifar}%
\end{table}%

\textbf{Results on CIFAR.}
We compare different training methods based on a ResNet-50 model with 400 training epochs. Besides the standard training method, we also compare our method with Dropout~\cite{srivastava2014dropout}.
By default, we use random cropping and horizontal flipping as the standard data augmentation.
Moreover, we consider state-of-the-art augmentation methods that are commonly used to improve robustness, such as AutoAugment~\cite{cubuk2018autoaugment} and AugMix~\cite{hendrycks2019augmix}.
Note that our EWS is complementary to these methods and does not require any modifications.

\begin{figure}[t]
  \centering
  \begin{subfigure}{0.48\linewidth}
    \includegraphics[width = 1.0\textwidth]{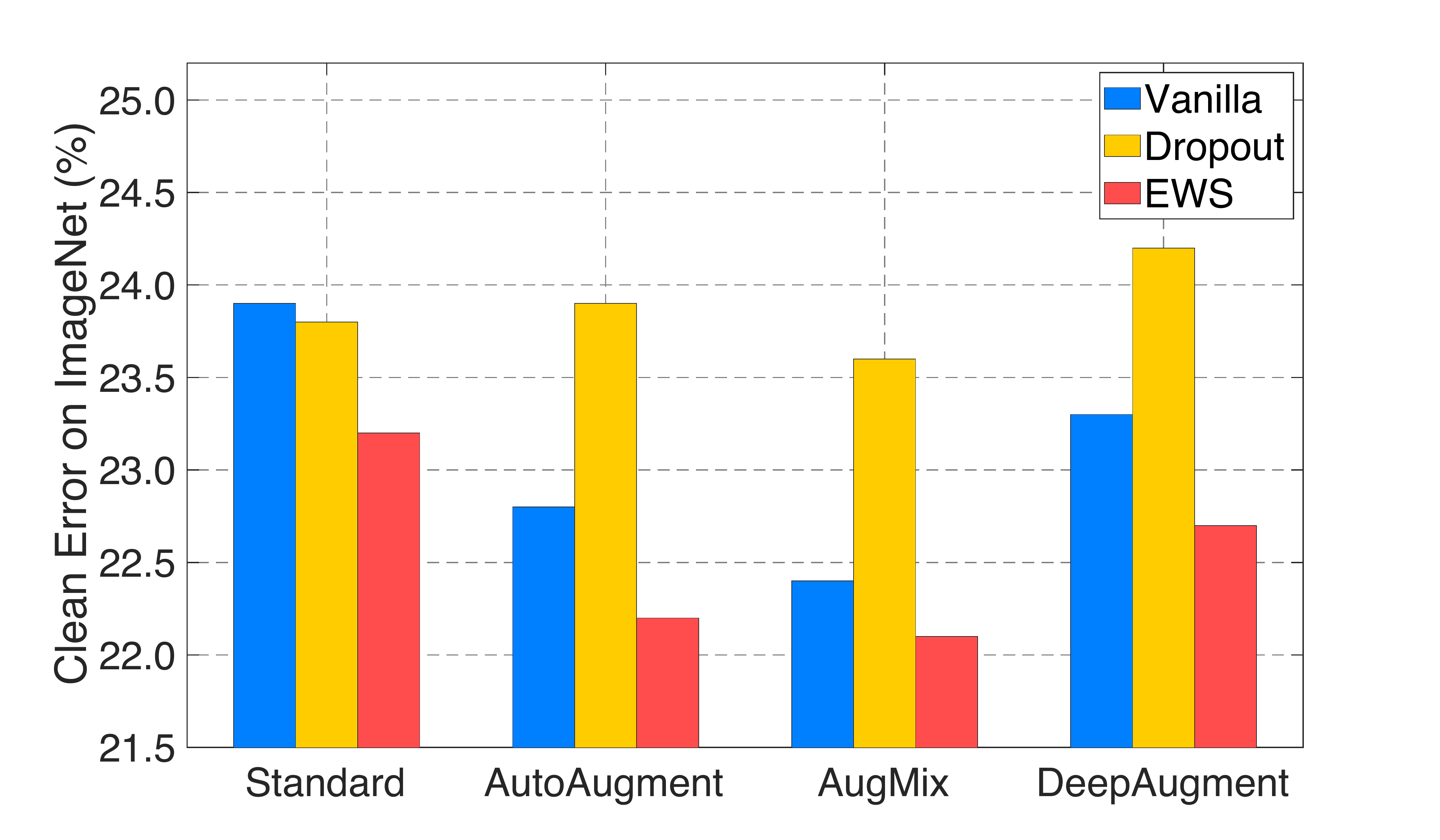} 
  \end{subfigure}
  \begin{subfigure}{0.49\linewidth}
    \includegraphics[width = 1.025\textwidth]{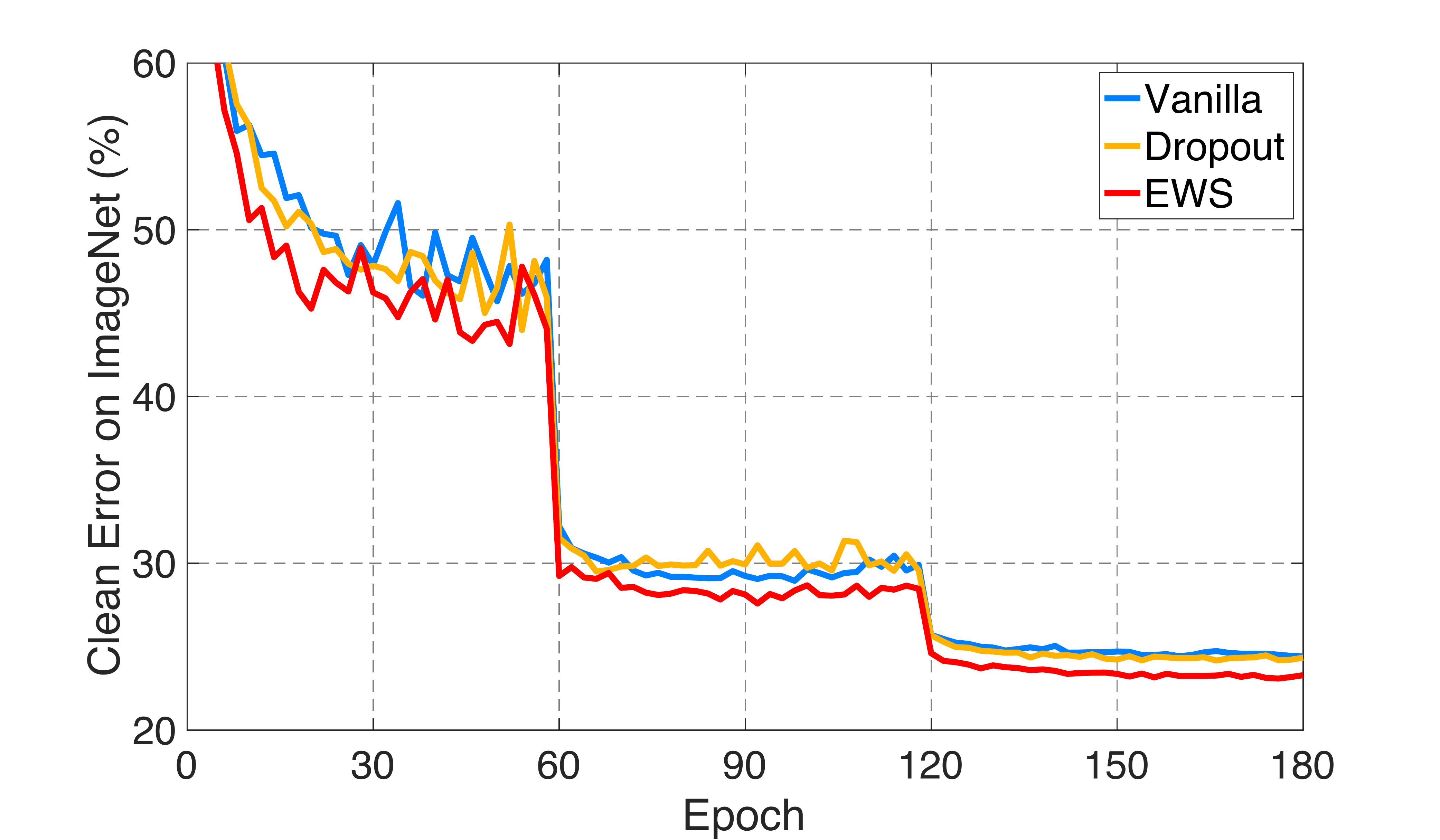} 
  \end{subfigure}
  \caption{
  \emph{Left:} Comparisons of clean top-1 test error on ImageNet. EWS consistently reduces error across diverse augmentation schemes. 
  \emph{Right:} Training curves in terms of top-1 test error on ImageNet using the standard data augmentation scheme. Clearly, the improvement of EWS can be observed throughout training.}
  \label{fig:imagenet}
\end{figure} 

{
Table~\ref{tab:cifar} shows that EWS is able to improve both clean and corrupted test errors over all three data augmentation schemes, \ie, ``standard'', AutoAugment and AugMix.
For example,
on CIFAR-10, we are able to improve the clean accuracy over AugMix by 0.59\% and significantly reduce the corruption error by 2.77\%.
Moreover, we also observe a similar performance improvement on CIFAR-100.
To be specific, our EWS reduces the corruption error by 3.04\% for AugMix while improving the clean accuracy by 0.64\% on CIFAR-100.
Note that the improvement is also significant for AutoAugment and the standard data augmentation, i.e., with the improvement of $1.88\%$ and $1.52\%$ on CIFAR-10, and the improvement of $1.97\%$ and $1.68\%$ on CIFAR-100.
We emphasize that the improvement gets more significant with more complex data augmentation.
This highlights that EWS is entirely complementary to state-of-the-art data augmentation.
Furthermore, our method outperforms Dropout in all settings. As Dropout can be interpreted as randomly selecting channels in the preceding layers, it is a natural baseline to compare our EWS against.
In addition, we also compare with a pruning based method CARDs~\cite{diffenderfer2021winning} that yields state-of-the-art results on CIFAR-10-C.
We highlight that, with AugMix, our EWS yields a larger relative improvement than CARDs~\cite{diffenderfer2021winning} (2.77\% vs 1.50\%, see more details in our supplementary).
}

\begin{table*}[t]
  \centering
  \resizebox{1.01\textwidth}{!}
  {
    \begin{tabular}{c|c|>{\columncolor{LightCyan}}c|ccccccccccccccc}
    \toprule
    \multicolumn{2}{c|}{\multirow{1}[0]{*}{Method}} & mCE $\downarrow$ & Gauss. & Shot & {Imp.} & Defoc. & Glass & Mot. & Zoom & Snow & Frost & Fog & Bright & Contra. & Elas. & Pixel & JPEG \\
    \hline
    \multirow{3}[0]{*}{Standard}  & Vanilla & 76.5 (-0.0)  & 80 & 82 & 83 & 75 & 89 & 78 & 80 & 78 & 75 & 66 & \textbf{57} & 71 & 85 & 77 & 77 \\
    & Dropout & 76.5 (-0.0) & 77 &	79 &	80 &	78 &	90 &	79 &	87 &	\textbf{77} &	77 &	67 &	58 &	\textbf{70} &	{84} &	75 &	76 \\
    & EWS & \textbf{75.1 (-1.4)} & \textbf{75} &	\textbf{76} &	\textbf{77} &	\textbf{73} &	\textbf{87} &	\textbf{77} &	\textbf{79} &	80 &	\textbf{73} &	\textbf{65} &	58 &	73 &	\textbf{83} &	\textbf{74} &	\textbf{75} \\
    \hline
    \multirow{2}[0]{*}{AutoAugment} 
    & Vanilla
     & 72.7 (-0.0) & 69 & {68} & 72 & \textbf{77} & {83} & 80 & 81 & 79 & 75 & {64} & 56 & {70} & 88 & 57 & \textbf{71} \\
     & EWS & \textbf{71.7 (-1.0)} & \textbf{67} & \textbf{68} &	\textbf{71} &	{78} &	\textbf{82} &	\textbf{78} &	\textbf{79} &	\textbf{78} &	\textbf{73} &	\textbf{64} &	\textbf{55} &	\textbf{69} &	\textbf{86} &	\textbf{56} &	{72}\\
     \hline
    \multirow{2}[0]{*}{AugMix} & Vanilla & 68.4 (-0.0) & 65 & 66 & 67 & {70} & \textbf{80} & 66 & 66 & 75 & 72 & 67 & 58 & \textbf{58} & {79} & 69 & \textbf{69} \\
    & EWS  & \textbf{67.5 (-0.9)} & \textbf{64} &	\textbf{63} & \textbf{63} &	\textbf{70} &	81 &	\textbf{65} &	\textbf{66} &	\textbf{72} &	\textbf{70} &	\textbf{64} &	\textbf{57} &	63 &	\textbf{79} &	\textbf{64} &	70 \\
    \hline
    \multirow{2}[0]{*}{DeepAugment} & Vanilla & 60.4 (-0.0) & 49 & 50 & 47 & 59 & 73 & 65 & 76 & 64 & \textbf{60} & 58 & 51 & 61 & 76 & 48 & 67 \\
    & EWS & \textbf{58.7 (-1.7)} & \textbf{48} & \textbf{48} &	\textbf{47} &	\textbf{58} &	\textbf{72} &	\textbf{58} &	\textbf{62} &	\textbf{63} &	{62} &	\textbf{58} &	\textbf{50} &	\textbf{56} &	\textbf{74} &	\textbf{47} &	\textbf{62}\\
    \toprule
    \end{tabular}%
    }
  \caption{Corruption error on ImageNet-C. We consider the mean corruption error (mCE) as well as the individual ones. In all data augmentation settings, EWS reduces mCE significantly.
  We just report Dropout for the standard setting as it usually worsens mCE on top of complex augmentation and put the full comparisons in our supplementary.
  }
  \label{tab:imagenetc}%
\end{table*}%

\begin{table*}[t]
  \centering
  \resizebox{0.87\textwidth}{!}
  {
    \begin{tabular}{c|c|>{\columncolor{LightCyan}}c|cccccccccc}
    \toprule
    \multicolumn{2}{c|}{\multirow{1}[0]{*}{Method}}& mFR $\downarrow$ & Gaussian & Shot & Motion & Zoom & Snow & Bright & Translate & Rotate & Tilt & Scale \\
    \hline
    \multirow{3}[0]{*}{Standard}  & Vanilla & 58.0 (-0.0) & \textbf{59} & 58 & 64 & 72 & 63 & 62 & \textbf{44} & 52 & 57 & \textbf{48} \\
    & Dropout & 57.8 (-0.2) & 62 & 59 &	65 &	52 &	48 &	58 &	63 &	57 &	{44} &	72 \\
    & EWS & \textbf{56.1 (-1.9)} & 62 & \textbf{55} &	\textbf{62} &	\textbf{49} &	\textbf{45} &	\textbf{52} &	64 &	\textbf{52} &	\textbf{42} &	71  \\
    \hline
    \multirow{2}[0]{*}{AutoAugment} & Vanilla
     & 51.7 (-0.0) & 50 & 45 & 57 & \textbf{68} & 63 & 53 & 40 & \textbf{44} & 50 & 46   \\
     & EWS & \textbf{50.4 (-1.3)} &  \textbf{48} &	\textbf{44} &	\textbf{53} &	70 &	\textbf{62} &	\textbf{52} &	\textbf{36} &	45 &	\textbf{49} &	\textbf{45}\\
     \hline
    \multirow{2}[0]{*}{AugMix} & Vanilla
     & 37.4 (-0.0) & 46 & 41 & \textbf{30} & 47 & 38 & 46 & \textbf{25} & \textbf{32} & 35 & 33 \\
    & EWS  & \textbf{36.6 (-0.8)} & \textbf{45} & \textbf{39} &	31 & \textbf{42} & \textbf{33} & \textbf{43} & 39 & 35 & \textbf{27} & \textbf{32}	 \\
    \hline
    \multirow{2}[0]{*}{DeepAugment} &  Vanilla  & 32.1 (-0.0) & 29 &	28 & 25 &	41  &	31 &	43 &	27 &	31 & 33 & 33 \\
    & EWS & \textbf{30.9 (-1.2)} & \textbf{28} & \textbf{26} & \textbf{25} & \textbf{40} & \textbf{28} & \textbf{41} & \textbf{26} & \textbf{30} & \textbf{32} & \textbf{33} \\
    \toprule
    \end{tabular}%
    }
  \caption{Mean flip rate (mFR) on ImageNet-P, testing stability of predictions on (corrupted) videos. In line with Table \ref{tab:imagenetc}, EWS improves consistently over all considered data augmentation schemes and nearly all corruption types. Similar to Table~\ref{tab:imagenetc}, Dropout did not improve over AutoAugment, AugMix or DeepAugment.}
  \label{tab:imagenetp}%
\end{table*}%

\textbf{Results on ImageNet.}
Again, we adopt a ResNet-50 as the baseline model. Following~\cite{hendrycks2019augmix}, we use the learning rate warm-up for the first 5 epochs and train the model for 180 epochs in total. In addition to AutoAugment and Augmix, we consider a stronger augmentation scheme DeepAugment~\cite{hendrycks2021many} designed for ImageNet.

In Fig.~\ref{fig:imagenet} (left), we first show that EWS consistently improves clean error on top of all considered data augmentation schemes. 
When equipped with AugMix, our EWS yields the best error of $22.1\%$ across all the considered settings.
The improvement of EWS can be also observed throughout the whole training process, as illustrated for the standard data augmentation scheme in Fig.~\ref{fig:imagenet} (right).
More critically, in Table \ref{tab:imagenetc} and \ref{tab:imagenetp} we focus on the significant robustness improvements on ImageNet-C and ImageNet-P through EWS.
Specifically, we consistently reduce mCE by $>0.9\%$ across different augmentation schemes on ImageNet-C.
We also highlight that EWS improves results across most included corruption types. 
Regarding ImageNet-P, these observations are further confirmed. For example, EWS reduces the mean flip rate (mFR) from $32.1\%$ to $30.9\%$ ($1.2\%$ improvement) on top of DeepAugment.
Overall, these results indicate that improving the performance of subnets through EWS boosts clean and corrupted error across a range of state-of-the-art data augmentation schemes.

\begin{table*}[t]
  \centering
  \resizebox{1.0\textwidth}{!}
  {
    \begin{tabular}{c|c|cc>{\columncolor{LightCyan}}c|cc>{\columncolor{LightCyan}}c|cc>{\columncolor{LightCyan}}c}
    \toprule
    \multicolumn{2}{c|}{\multirow{2}[0]{*}{Method}} & \multicolumn{3}{c|}{PreAct ResNet-18} & \multicolumn{3}{c|}{WRN-28-10} & \multicolumn{3}{c}{WRN-34-10} \\ 
    \multicolumn{1}{c}{} &  & {Clean $\downarrow$} & {PGD-20 $\downarrow$} & {AA $\downarrow$} & {Clean $\downarrow$} & {PGD-20 $\downarrow$} & {AA $\downarrow$} & {Clean $\downarrow$} & {PGD-20 $\downarrow$} & {AA $\downarrow$} \\
    \hline
    \multirow{4}[0]{*}{AT} & Vanilla~\cite{madry2017towards} &   17.54	    &   49.18   & 52.96 (-0.00) & 14.89	& 45.17 & 47.81 (-0.00)  & 14.74  & 45.39 & 47.47 (-0.00) \\
          & EWS   &   \textbf{16.85}	    & \textbf{47.99}     & \textbf{51.84 (-1.12)} & \textbf{14.57} &	\textbf{44.24} & \textbf{47.17 (-0.64)} & \textbf{14.33}	& \textbf{44.04} & \textbf{46.58 (-0.89)} \\
          \cline{2-11}
          & AWP~\cite{wu2020adversarial}   &  19.59	    &  46.01    & 51.43 (-0.00) & 15.89 & 42.93 & 46.41 (-0.00) &  \textbf{14.17} & 41.89   & 45.96 (-0.00) \\
          & AWP-EWS &  \textbf{19.25}	    & \textbf{44.98}      & \textbf{50.48 (-0.95)} & \textbf{15.81} &	\textbf{41.72} & \textbf{45.58 (-0.83)} & 14.21	& \textbf{41.07} & \textbf{45.29 (-0.67)} \\
    \hline
    \multirow{4}[0]{*}{TRADES} & Vanilla~\cite{zhang2019theoretically} & 17.42   &  46.88    & 50.84 (-0.00) & 15.50 & 44.11 & 47.40 (-0.00) & 15.32 & 43.84 & 46.89 (-0.00) \\
          & EWS   &  \textbf{17.10}    &  \textbf{45.73}    & \textbf{49.67 (-1.17)} & \textbf{15.09} & \textbf{43.45} & \textbf{46.72 (-0.68)}  & \textbf{14.56} & \textbf{43.13} & \textbf{46.06 (-0.83)} \\
          \cline{2-11}
          & AWP~\cite{wu2020adversarial}   &  18.27	    &  45.36   &  49.62 (-0.00) & 14.84 & 41.25 & 44.86 (-0.00) & 15.55 & 40.85 & 43.90 (-0.00) \\
          & AWP-EWS &   \textbf{17.67}    &  \textbf{44.20}    & \textbf{48.58 (-1.04)} & \textbf{14.30}	& \textbf{40.40} & \textbf{44.22 (-0.64)}  & \textbf{14.13} & \textbf{40.05} & \textbf{43.17 (-0.73)} \\
         \hline \multicolumn{2}{c|}{TRADES-AWP*} & 17.13 & 43.68 & 48.37 (-0.00) & 13.37 & 38.51 & 41.97 (-0.00) & 12.73 & 35.97 & 40.74 (-0.00) \\
         \multicolumn{2}{c|}{TRADES-AWP-EWS*} & \textbf{16.62} & \textbf{42.33} & \textbf{47.23 (-1.14)} & \textbf{12.59} & \textbf{37.60} & \textbf{41.23 (-0.74)}  & \textbf{11.90} & \textbf{35.19} & \textbf{40.05 (-0.69)}\\
    \bottomrule
    \end{tabular}%
    }
  \caption{Clean and robust test error, \ie, on adversarial examples generated using PGD-20 and AutoAttack, for three different architectures: PreAct ResNet-18, WRN-28-10 and WRN-34-10. 
  We consider `vanilla'' AT and TRADES as well as their AWP variants as the baselines. We also highlight the improvement against AutoAttack in parentheses. * denotes the models trained with additional 500K unlabeled data~\cite{CarmonNIPS2019}.
  Across all the settings, EWS reduces both the robust test error and the clean error.
  }
  \label{tab:adv_resnet}%
\end{table*}%

\subsection{Improving Adversarial Robustness}
\label{subsec:experiments-adversarial}

Besides corruption robustness, on CIFAR-10, we also apply EWS on top of several popular adversarial training approaches, including vanilla adversarial training (AT)~\cite{madry2017towards},  TRADES~\cite{zhang2019theoretically}, AT with adversarial weight perturbations (AT-AWP)~\cite{wu2020adversarial}, and using additional unlabeled examples~\cite{CarmonNIPS2019}.
Our setup follows the settings of~\cite{RiceICML2020}.
For AWP, we follow the hyper-parameters of the original paper \cite{wu2020adversarial}.
Note that combinations with these approaches are also possible, \eg, TRADES with AWP \emph{and} EWS, highlighting the flexibility of EWS.
We consider PreAct ResNet-18 \cite{he2016identity}, WRN-28-10 and WRN-34-10 \cite{zagoruyko2016wide} and employ early stopping \cite{RiceICML2020}.
We train and evaluate using an $\epsilon$ of $\nicefrac{8}{255}$. During training we use $10$ iterations of PGD.
At test time, we evaluate models under AutoAttack (AA)~\cite{CroceICML2020} and PGD with $20$ iterations.

{
In Table \ref{tab:adv_resnet}, our EWS consistently yield significant improvement in term of adversarial robustness across both architectures and adversarial training variants.
For example, considering vanilla AT, EWS reduces clean error of PreAct ResNet-18 by 1.19\% and robust error against AA by 1.12\%.
While the improvement gets slightly smaller for larger models, \ie, WRN-28-10 and WRN-34-10, EWS improves both clean and robust error consistently.
On a WRN-34-10, the improvement in robust error is still $0.89\%$. 
We highlight that we can yield a similar improvement of 0.77\% against AA when we consider stronger training tricks specified by~\cite{PangICLR2021}, i.e., additionaly using label smoothing and Softplus (see more details in the supplementary).
This improvement also generalizes to TRADES and AWP both of which generally improve adversarial robustness.
Moreover, EWS is able to improve over TRADES and AWP when using additional pseudo-labeled training examples, which performs best in our experiments.
Here, EWS reduces the robust test error from $40.74\%$ to $40.05\%$ ($0.69\%$ improvement).
At the same time, EWS also reduces the clean error by $0.83\%$.
EWS improving over AWP also indicates that utilizing adversarial weight perturbations \cite{wu2020adversarial} \emph{on} weak subnets instead of the overall network is more beneficial.
We also compare our method with a very recent adversarial training method LBGAT~\cite{cui2021learnable}.
We show that, using the same settings of LBGAT~\cite{cui2021learnable}, EWS yields better adversarial robustness and clean accuracy than LBGAT on both CIFAR-10 and CIFAR-100. We put the detailed comparisons in supplementary due to page limit. 
Overall, these experiments highlight the generality of our method, improving not only corruption robustness but also adversarial robustness.}

\begin{table}[t]
  \centering
  \resizebox{0.63\textwidth}{!}
  {
    \begin{tabular}{c|c|c}
    \toprule
    Method & \multicolumn{1}{c|}{~Clean Error (\%) $\downarrow$~} & \multicolumn{1}{c}{~Corruption Error (\%) $\downarrow$~} \\
    \hline
    Baseline & 5.32  & 26.46 \\
    \hline
    Random Search & 4.49  & 25.81 \\
    ~$L_1$-norm Selection~ & 4.37  & 25.45 \\
    Ours   & \textbf{4.12}  & \textbf{24.94} \\
    \bottomrule
    \end{tabular}%
    }
   \caption{Clean and corrupted test error on CIFAR-10 and CIFAR-10-C.
  We compare our search method with random search and an $L_1$-norm based heuristic method. While randomly selecting subnets to improve already reduces both clean and corrupted error, finding particularly weak subnets obtains the lowest errors.}
    \label{tab:comparison_search}%
\end{table}%

\section{Ablation and Discussions}
\label{sec:ablation}

In the following, we present further ablation experiments and discussions.
Specifically, in Section~\ref{subsec:ablation}, we demonstrate that finding particularly weak subnets is important and discuss training cost as well as hyper-parameters.
This also shows that computational overhead is minimal, even for adversarial training.
Furthermore, we study the weight of the distillation loss $\lambda$ and the width of the selected subnets $\rho$.
Finally, in Section~\ref{subsec:analysis}, we show that EWS allows to analyze the vulnerability of individual blocks and layers.

\subsection{Search Strategies and Hyper-Parameters}
\label{subsec:ablation}

We perform ablations on CIFAR-10(-C) regarding our search strategy and hyper-parameters. For all the experiments, we use a ResNet-50 and adopt the same settings as before.

\noindent\textbf{Search strategies.}
In Table \ref{tab:comparison_search}, we conduct an ablation study to investigate the effectiveness of our search method.
We compare our method to two baselines: random search selects subnets entirely at random; and $L_1$-norm selection chooses the channels/paths with lowest $L_1$-norm of weights. For simplicity, we compare clean and corrupted test error on CIFAR-10(-C). Since the randomly sampled subnets may contain some weak components, we are able to reduce both the clean and corrupted error. Using $L_1$-norm is a simple heuristic to find weaker subnets than random search. This can be seen by a slightly reduced clean and corrupted error. 
Nevertheless, the $L_1$-norm may not be highly correlated with accuracy and the search results are often suboptimal. In contrast, our method yields the lowest clean and corrupted errors as the controller directly finds subnets with low accuracy. This also confirms our results in Fig.~\ref{fig:motivation}.

\begin{figure}[t]
  \centering
    \includegraphics[width = 0.58\textwidth]{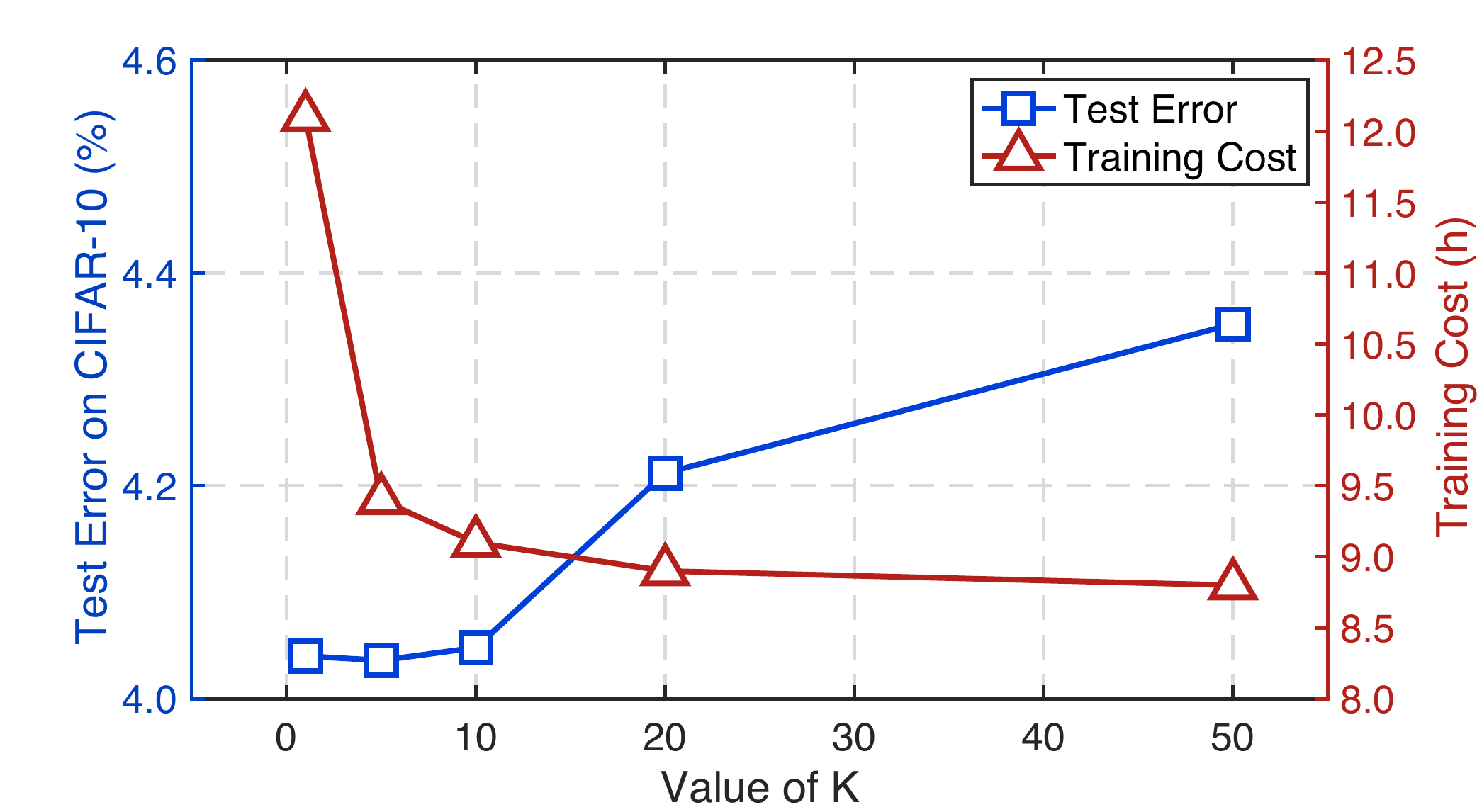}
    \caption{We plot (clean) test error on CIFAR-10 ({\color{blue}blue}) and training cost in hours ({\color{red}red}) against the controller training interval $K$, see Algorithm \ref{alg:training}. Clearly, increasing $K$ reduces training cost significantly. 
    At the same time, clean error reduces significantly for $K \leq 10$. 
    In practice, $K = 10$ yields a good trade-off and only introduces little computational overhead (roughly 3\%) for both corruption and adversarial experiments.
    }
    \label{fig:ablation_k}
\end{figure}

\noindent\textbf{Training interval $K$ and training cost.}
As detailed in Section~\ref{subsec:training} and Algorithm~\ref{alg:training}, we train the controller model every $K$ iterations.
Thus, we expect a trade-off between training cost and performance: a stronger controller model should improve performance, but needs to be updated more often (\ie, lower $K$) which, however, increases the training cost.
As shown in Fig.~\ref{fig:ablation_k}, gradually increasing $K$ greatly reduces the training cost. Note that we obtain promising results for $K {\leq} 10$ but observe a significant increase of test error when $K {>} 10$.
We observe that $K{=}10$ also works well for ImageNet(-C) and the data dependent tuning is not necessary.
This justifies our choice of $K{=}10$ which also results in small computational overhead of roughly $3\%$ on the adversarial training settings.

\begin{figure}[t]
  \centering
  \begin{subfigure}{0.49\linewidth}
    \includegraphics[width = 1.0\textwidth]{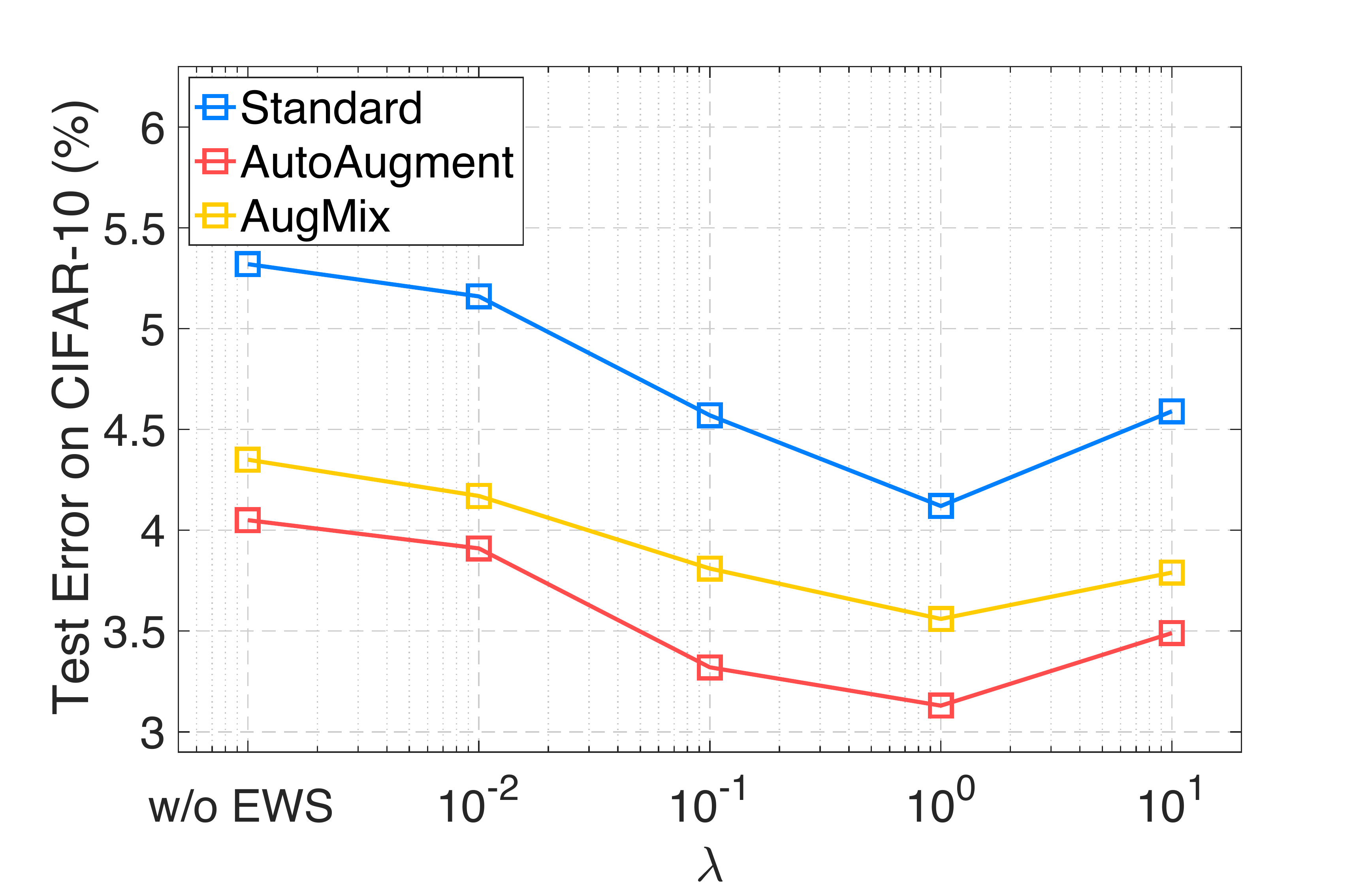} 
  \end{subfigure}
  \begin{subfigure}{0.49\linewidth}
    \includegraphics[width = 1.0\textwidth]{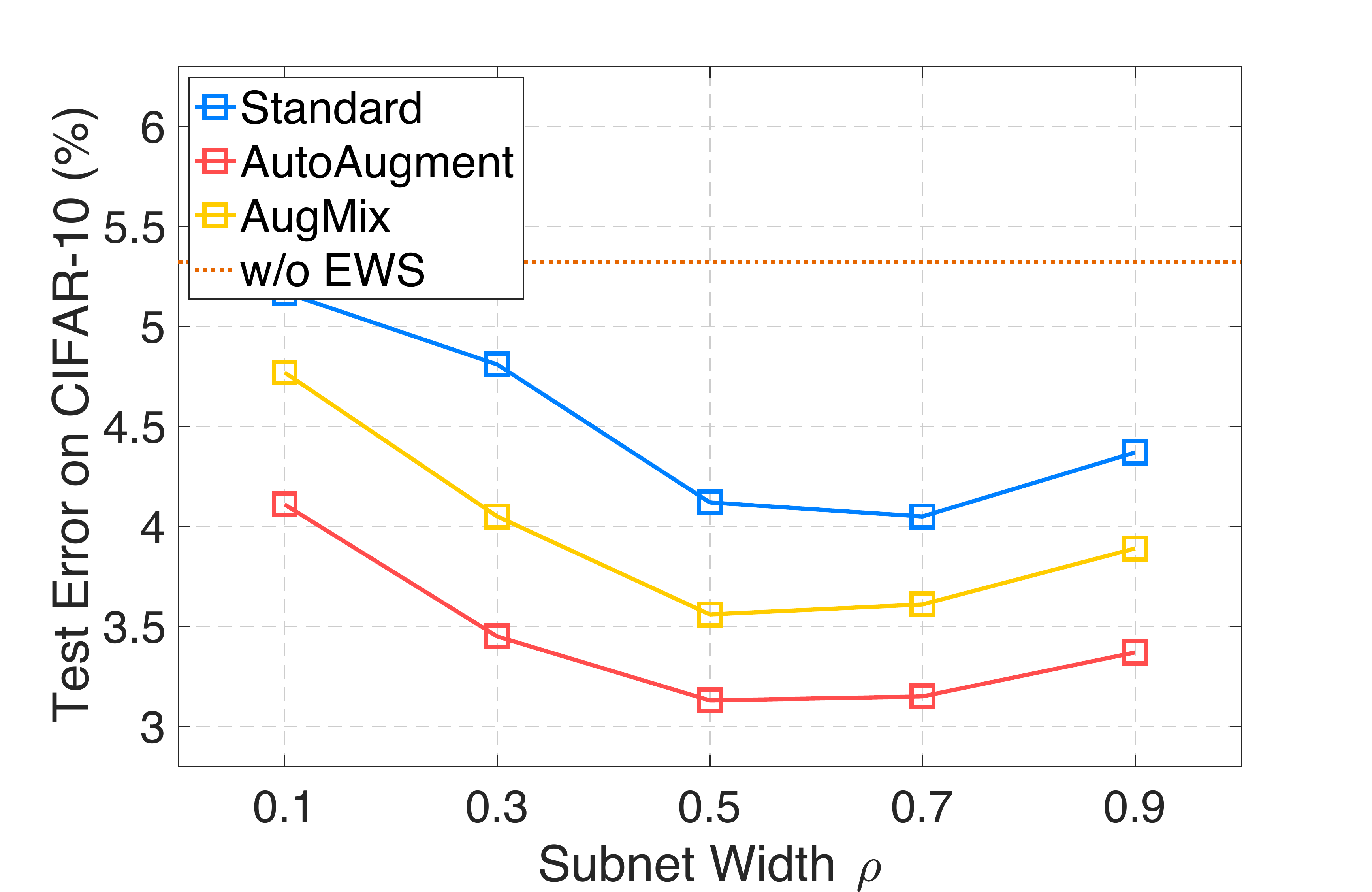} 
  \end{subfigure}
  \caption{Clean test error on CIFAR-10 plotted against the weight of the distillation loss (left, see Algorithm \ref{alg:training}) and the subnet width $\rho$ (right). \emph{Left:} Across all tested data augmentation schemes, including AutoAugment and AugMix, $\lambda=1$ performs best. \emph{Right:} Too small or too large subnets during training reduce the benefit of EWS. We found that $\rho = 0.5$ and $\rho = 0.7$ perform best in most cases.
  }
  \label{fig:lambda_width}
\end{figure}

\noindent\textbf{Weight of distillation loss.}
In Fig.~\ref{fig:lambda_width} (left), we change the value of $\lambda$ in Eqn.~\eqref{eq:ews_clean}.
Note that $\lambda=0$ corresponds to training \emph{without} EWS.
Larger $\lambda$, in contrast, increases the effect of EWS, \ie, trying to enhance weak subnets more aggressively.
Given a set of values $\lambda \in \{0, 0.001, 0.01, 0.1, 1, 10\}$, we gradually reduce clean test error up until $\lambda = 1$. Larger values will result in a small performance degradation. However, even $\lambda = 10$ still outperforms training without EWS. This generalizes across all considered data augmentation schemes, including AutoAugment and AugMix.

\noindent\textbf{Subnet width.}
In Fig.~\ref{fig:lambda_width} (right), we investigate the impact of subnet width $\rho$ on the performance improvement obtained using EWS. To this end, we consider a set of candidate widths $\rho \in \{0.1, 0.3, 0.5, 0.7, 0.9\}$. We suspect that very small subnets ($\rho{=}0.1$) cause the distillation loss $\mL_{\rm KL}\big(\alpha(x), M(x) \big)$ to be very large, hampering the training process. When we increase $\rho$ to $0.5$ or $0.7$, EWS yields significantly better results. For a larger $\rho \geq 0.9$, the distillation loss can be very small, performance drops again.

\subsection{Vulnerability of Blocks and Layers}
\label{subsec:analysis}
\begin{figure}[t]
  \centering
    \includegraphics[width = 0.73\textwidth]{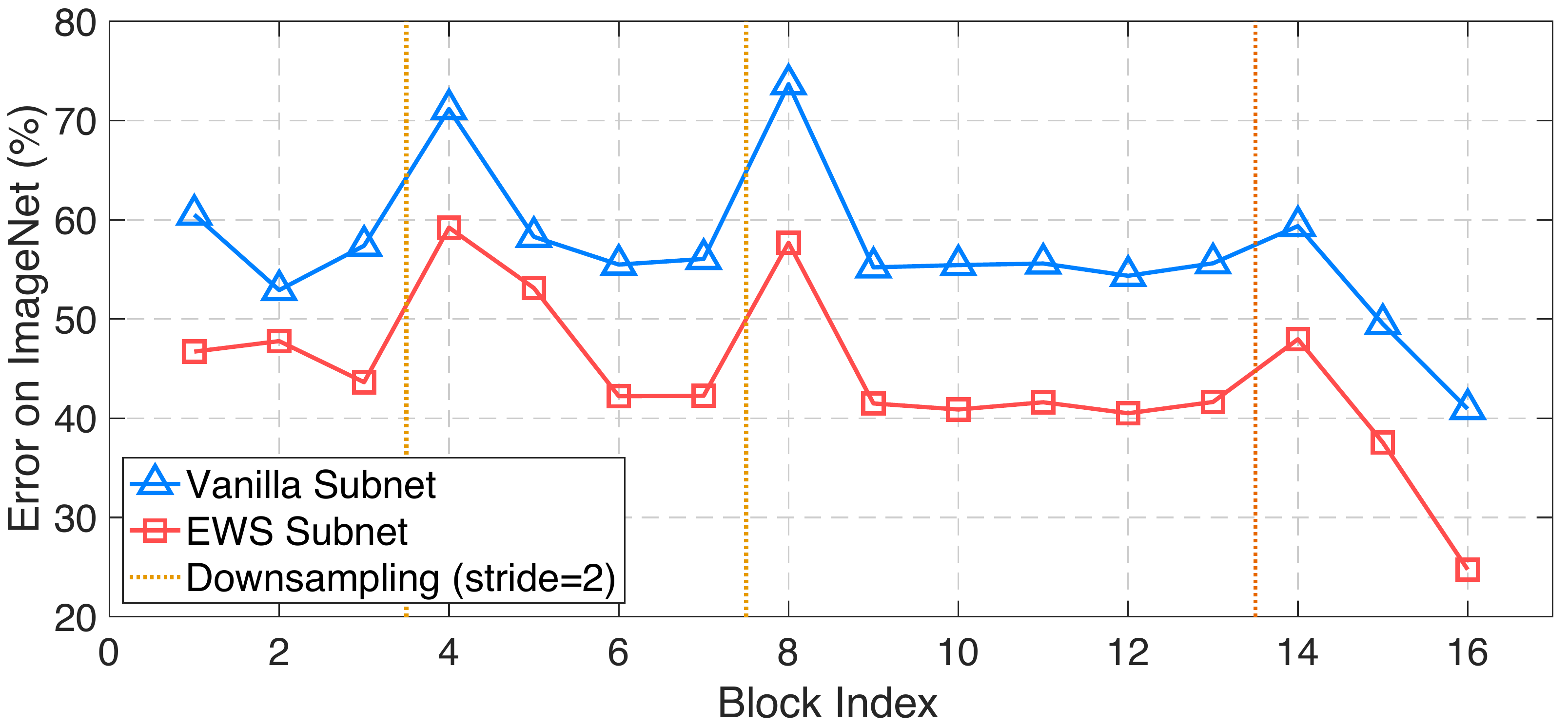} 
    \caption{
    Test error on ImageNet when constructing subnets by randomly selecting $50\%$ of paths/channels in a specific block while keeping the rest unchanged. The results are averaged across $100$ sampled subnets. The \textcolor{orange}{dotted line} marks the position of downsampling operation. Clearly, blocks \emph{after} downsampling blocks are most vulnerable and the last two blocks are least vulnerable. In all cases, EWS reduces error on subnets significantly. 
    }
    \label{fig:which_layer}
\end{figure}

As mentioned in Section~\ref{subsec:subnets}, we construct subnets by selecting a subset of paths and channels.
This also allows us to investigate the vulnerability of each block. To this end, we construct subnets by randomly selecting a subset of paths/channels in a specific block, while keeping all other blocks unchanged. In Fig.~\ref{fig:which_layer}, we report test error on ImageNet when randomly sampling 100 subnets for each block. 
We observed that the blocks behind the downsampling operation (\textcolor{orange}{dotted lines}) tend to be more vulnerable. More critically, EWS consistently reduces test error across all blocks compared to standard training.

\section{Conclusion}

In this paper, we described the phenomenon that most subnets of deep networks perform rather poorly, especially on perturbed examples. Interestingly, these weak subnets are highly correlated with the lack of robustness of the full model.
To address this issue, we focused on improving model robustness by identifying and enhancing these particularly weak subnets.
This leads to our proposed training method, EWS, which specifically employs a search algorithm to find weak subnets and then strengthens them through knowledge distillation.
With minimal computational overhead, EWS can be applied on top of popular data augmentation schemes as well as adversarial training variants.
Experiments show that EWS improves robustness consistently over these methods.

\clearpage
%
%
\bibliographystyle{splncs04}
\bibliography{egbib}
\end{document}

%% file: def.tex
\def\mD{{\mathcal D}}

\def\mL{{\mathcal L}}

\def\0{{\bf 0}}
\def\1{{\bf 1}}

\def\eg{{e.g.}} 
\def\ie{{i.e.}}